\documentclass[sigconf]{acmart}
\usepackage{graphicx}
\usepackage{amsthm}

\usepackage{amssymb}
\usepackage{algorithm}
\usepackage{algorithmic}
\usepackage{multirow}
\usepackage{stfloats}

\usepackage{amsmath,amsfonts,bm}

\def\eqref#1{equation~\ref{#1}}

\def\1{\bm{1}}

\DeclareMathAlphabet{\mathsfit}{\encodingdefault}{\sfdefault}{m}{sl}
\SetMathAlphabet{\mathsfit}{bold}{\encodingdefault}{\sfdefault}{bx}{n}

\usepackage{hyperref}
\usepackage{url}
\AtBeginDocument{%
  \providecommand\BibTeX{{%
    \normalfont B\kern-0.5em{\scshape i\kern-0.25em b}\kern-0.8em\TeX}}}

\setcopyright{acmcopyright}

\copyrightyear{2024}
\acmYear{2024}
\setcopyright{acmlicensed}\acmConference[KDD '24]{Proceedings of the 30th ACM SIGKDD Conference on Knowledge Discovery and Data Mining}{August 25--29, 2024}{Barcelona, Spain}
\acmBooktitle{Proceedings of the 30th ACM SIGKDD Conference on Knowledge Discovery and Data Mining (KDD '24), August 25--29, 2024, Barcelona, Spain}
\acmDOI{10.1145/3637528.3671965}
\acmISBN{979-8-4007-0490-1/24/08}

\acmConference[KDD'24]{the 30th ACM SIGKDD Conference on Knowledge Discovery and Data Mining}{August 25--28,
  2024}{Barcelona, Spain}
\acmPrice{15.00}
\acmISBN{978-1-4503-XXXX-X/18/06}

\begin{document}

\title{Large Language Model-driven Meta-structure Discovery in Heterogeneous Information Network}



\author{Lin Chen}
\affiliation{
 \institution{Hong Kong University of Science and Technology, Hong Kong, China}
   \city{} 
   \country{}} 
\email{lchencu@connect.ust.hk}

\author{Fengli Xu}
\affiliation{
 \institution{Department of Electronic Engineering, BNRist,  Tsinghua University, Beijing, China}
   \city{} 
   \country{}} 
\email{fenglixu@tsinghua.edu.cn}
\authornote{Fengli Xu, Yong Li, and Pan Hui are the corresponding authors.}

\author{Nian Li}
\affiliation{
 \institution{Shenzhen International Graduate School, Tsinghua University, Shenzhen, China}
   \city{} 
   \country{}} 
\email{linian21@mails.tsinghua.edu.cn}

\author{Zhenyu Han}
\affiliation{
 \institution{Department of Electronic Engineering, BNRist, Tsinghua University, Beijing, China}
   \city{} 
   \country{}} 
\email{hanzy19@mails.tsinghua.edu.cn}

\author{Meng Wang}
\affiliation{
 \institution{Hefei University of Technology, Hefei, China} 
   \city{} 
   \country{}} 
\email{wangmeng@hfut.edu.cn}
   
\author{Yong Li}
\affiliation{
 \institution{Department of Electronic Engineering, BNRist, Tsinghua University, Beijing, China}
   \city{} 
   \country{}} 
\email{liyong07@tsinghua.edu.cn}
\authornotemark[1]

\author{Pan Hui}
\affiliation{
 \institution{Hong Kong University of Science and Technology (Guangzhou), China}
   \city{} 
   \country{} 
 \institution{Hong Kong University of Science and Technology, Hong Kong, China}
    \city{}
    \country{}
}
\email{panhui@ust.hk}
\authornotemark[1]

\renewcommand{\shortauthors}{Chen et al.}

\begin{abstract}
Heterogeneous information networks (HIN) have gained increasing popularity in recent years for capturing complex relations between diverse types of nodes. 
Meta-structures are proposed as a useful tool to identify the important patterns in HINs, but hand-crafted meta-structures pose significant challenges for scaling up, drawing wide research attention towards developing automatic search algorithms.
Previous efforts primarily focused on searching for meta-structures with good empirical performance, overlooking the importance of human comprehensibility and generalizability. 
To address this challenge, we draw inspiration from the emergent reasoning abilities of large language models (LLMs).
We propose \emph{ReStruct}, a meta-structure search framework that integrates LLM reasoning into the evolutionary procedure.
\emph{ReStruct} uses a \emph{grammar translator} to encode the meta-structures into natural language sentences, and leverages the reasoning power of LLMs to evaluate their semantic feasibility. 
Besides, \emph{ReStruct} also employs performance-oriented evolutionary operations. 
These two competing forces allow \emph{ReStruct} to jointly optimize the semantic explainability and empirical performance of meta-structures.
Furthermore, \emph{ReStruct} contains a \emph{differential LLM explainer} to generate and refine natural language explanations for the discovered meta-structures by reasoning through the search history.
Experiments on eight representative HIN datasets demonstrate that \emph{ReStruct} achieves state-of-the-art performance in both recommendation and node classification tasks. 
Moreover, a survey study involving $73$ graduate students shows that the discovered meta-structures and generated explanations by \emph{ReStruct} are substantially more comprehensible.  
Our code and questionnaire are available at \url{https://github.com/LinChen-65/ReStruct}.
\end{abstract}




\maketitle
\section{Introduction}

Heterogeneous information networks (HINs) are effective in jointly modeling network topology and multi-typed relations~\cite{shi2016survey}, leading to their widespread adoption across various applications, such as social media~\cite{xu2019relation}, information retrieval~\cite{wang2019heterogeneous}, and recommender systems~\cite{han2020genetic,ding2024artificial}. 
To fully exploit the rich semantic information encoded in HINs, researchers have proposed to use \emph{meta-path}s, which are templates of relation sequences to model the complex proximity on HINs~\cite{sun2011pathsim}. 
They were later extended to \emph{meta-structure}s to capture more general interaction patterns beyond linear paths~\cite{huang2016meta}. 
These meta-structures have been successfully utilized in heterogeneous graph neural networks (GNNs) to learn expressive representations for HINs~\cite{xu2019relation,wang2019heterogeneous} for completing downstream tasks. 
However, the reliance on hand-crafted meta-structures, which depend on domain experts' knowledge, makes it challenging to scale up to larger and more complex HINs that are commonly encountered for real-world applications.

Driven by the importance of domain adaptation, recent research efforts have been dedicated to developing algorithms for automatic meta-structure search. 
Researchers propose to use genetic algorithm~\cite{han2020genetic}, deep reinforcement learning~\cite{peng2021reinforced} and differentiable neural architectural search models~\cite{ding2021diffmg} to automatically identify meta-structures that can enhance the performance of heterogeneous GNNs. 
However, these previous attempts primarily focus on the prediction performance of meta-structures, often resulting in highly complex structures that are challenging to interpret and prone to overfitting. 
Such ``meta-structures'' deviate from the original inspiration of meta-structure research that aims to extract semantically clear features from HINs~\cite{huang2016meta}.

The recent breakthrough in large language models (LLMs)~\cite{floridi2020gpt} offers a unique opportunity to tackle the challenges of meta-structure discovery. 
The scaled-up versions of LLMs have exhibited emergent abilities for a wide range of complex tasks that go beyond auto-regression token generation~\cite{wei2022emergent}. 
For example, researchers have found that chain-of-thoughts prompting can effectively unlock LLMs' reasoning capability for commonsense, mathematical, and logical problems~\cite{wei2022chain}. 
Such a general-purpose reasoning capability holds huge potential for comprehending the rich semantic information and produce human understandable knowledge from given HINs, which could be path-breaking to current performance-oriented meta-structure search algorithms. 

In this paper, we propose a novel framework named \emph{ReStruct} (short for \underline{\textbf{Re}}asoning meta-\underline{\textbf{Struct}}ure search) that integrates LLM reasoning into an evolutionary procedure for meta-structure search. 
In this framework, we design a \emph{grammar translator} to encode meta-structures into natural language sentences with nested clauses (see Figure~\ref{fig:grammar}), ensuring that their semantic meanings can be readily comprehended by LLMs. 
Besides, we define a set of basic operations to modify a given meta-structure, allowing \emph{ReStruct} to explore its adjacent possibilities in a valid space.
Unlike pure performance-oriented search, we anticipate \emph{ReStruct} to evaluate both semantic feasibility and empirical performance to identify promising candidates. 
To this end, we first design a \emph{few-shot LLM predictor} to estimate the performance of meta-structure candidates with access to previously evaluated meta-structures from a history pool, followed by a \emph{similarity-oriented LLM selector} to identify the most promising candidates based on the semantic similarities.   
After empirically evaluating the chosen candidates with heterogeneous GNNs, we design an \emph{evolutionary updater} adopting the classic \emph{elimination-reproduction} procedure to refine meta-structure candidates based on their performances. 
Finally, we design a \emph{differential LLM explainer} that generates natural language explanations for the discovered meta-structure. 
It employs a chain-of-thought prompting technique to perform step-by-step \emph{structural comprehension} and \emph{performance attribution}. 
This reasoning process generates high-quality explanations by explicitly comparing the chosen meta-structures and the adjacent yet unchosen ones. 

We evaluate \emph{ReStruct} on eight representative HIN datasets. 
Experiments show that \emph{ReStruct} achieves state-of-the-art performance on both recommendation and node classification tasks, and generates meaningful explanations as it searches through the solution space. 
To effectively assess the explainability of the discovered meta-structures, we conduct a user survey on $73$ graduate students with domain knowledge in HIN research.
According to the survey results, 46.6\% of the participants consider the meta-structure discovered by \emph{ReStruct} as the most comprehensible compared with three strong baselines, outperforming the second best baseline by 61.8\%. 
Moreover, the natural language explanations generated by \emph{ReStruct} are significantly preferred by the majority (77.6\% on average) in a head-to-head comparison with baseline methods.

We summarize our main contributions below: 
\begin{itemize}
    \item We propose a novel \emph{ReStruct} framework that integrates LLM reasoning into an evolutionary meta-structure search procedure. \emph{ReStruct} jointly optimizes the empirical prediction performance and semantic explainability of meta-structures, by coordinating the competing forces of an \emph{evolutionary updater} and a \emph{semantic similarity-oriented LLM selector}. 
    This represents a significant advancement in meta-structure search algorithms, enabling the generation of meta-structures that represent human digestible knowledge on HINs and are less prone to overfitting. 
    \item We design a \emph{grammar translator} to encode meta-structures as natural language sentences, which unleashes the reasoning power of LLMs to make sense of the rich semantic information on HINs. 
    On top of this, we design a \emph{differential LLM explainer} that can generate human-comprehensible natural language explanations for discovered meta-structures.
    \item We conduct extensive experiments to reveal \emph{ReStruct}'s state-of-the-art performance on eight representative datasets. Furthermore, we carry out a user survey to validate that \emph{ReStruct} substantially outperforms baseline methods in terms of the comprehensibility of discovered meta-structures and usefulness of generated explanations.
\end{itemize}
\section{Preliminaries}

Here, we provide the definitions of heterogeneous information networks, meta-paths, and meta-structures as in the literature.

\textit{Definition 2.1.} \textbf{Heterogeneous Information Network (HIN)}~\cite{sun2011pathsim}. 
An information network (IN) is mathematically a graph denoted as $G = \{V, E, T, R, \sigma, \phi\}$, with $V = \{v_1, v_2, ..., v_n\}$ being the set of nodes, $E = \{e_1, e_2, ..., e_m\}$ being the set of edges, $T = \{t_1, t_2, ...,t_k\}$ being the set of node types, and $R = \{r_1, r_2, ..., r_j\}$ being the set of edge types. 
$\sigma: V \rightarrow T$ is a function that maps each node to its associated type, and $\phi: E \rightarrow R$ is a function that maps each edge to its associated type.
The network schema of $G$ is then denoted as $S=\{T, R\}$.
If $|T|>1$ (multiple types of nodes) or $|R|>1$ (multiple types of edges), $G$ is a heterogeneous information network (HIN). 
Otherwise, it is a homogeneous information network.

\textit{Definition 2.2.} \textbf{Meta-path}~\cite{sun2011pathsim}.
Given an HIN $G$, a meta-path $P = t_1 \xrightarrow{e_1} t_2 ... \xrightarrow{e_{p-1}} t_p$, is a sequence of node types and edge types defined on the network schema $S$, connecting a single source node type and a single target node type. 
One meta-path may correspond to many meta-path instances in $G$.

\textit{Definition 3.3.} \textbf{Meta-structure}~\cite{huang2016meta}.
Given an HIN $G$, a meta-structure $T$ is a generalization of the meta-path to allow for the existence of graph structures beyond linear connections between the source node type and the target node type.
\section{Methods}

\begin{figure*}[ht]
    \centering
    \includegraphics[width=0.95\linewidth]{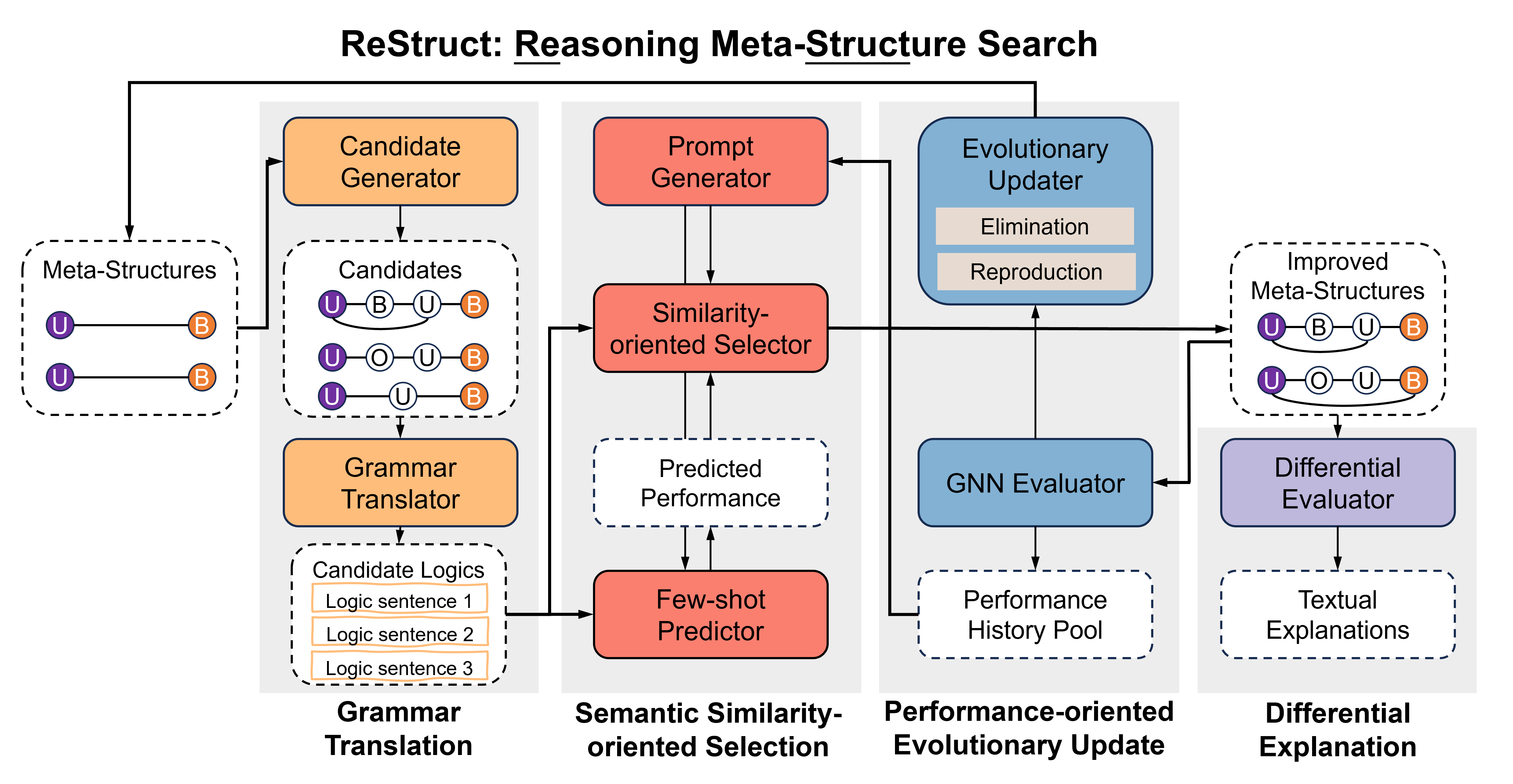}
    \caption{Overview of our proposed \emph{ReStruct} framework.}
    \Description{Overview of our proposed \emph{ReStruct} framework.}
    \label{fig:framework}
\end{figure*}

In this section, we provide a detailed introduction to our proposed methods.
In Section~\ref{grammar}, we elaborate our novel design of natural language encoding of meta-structures to facilitate LLMs' understanding of its semantic meanings. 
In Section~\ref{method:basic_operations}, we introduce our design of three basic operations for generating candidate meta-structures.
In Section~\ref{method:llm_selector}, we design two LLM agents to evaluate and select candidate meta-structures with semantic similarity orientation.
In Section~\ref{framework}, we combine LLM-guided optimization with evolutionary processes to form an effective derivative-free optimization framework.
The overview of our framework is shown in Figure~\ref{fig:framework}.

\subsection{Natural Language Encoding of Meta-Structures} \label{grammar}

\begin{figure}[ht]
    \centering
    \includegraphics[width=\linewidth]{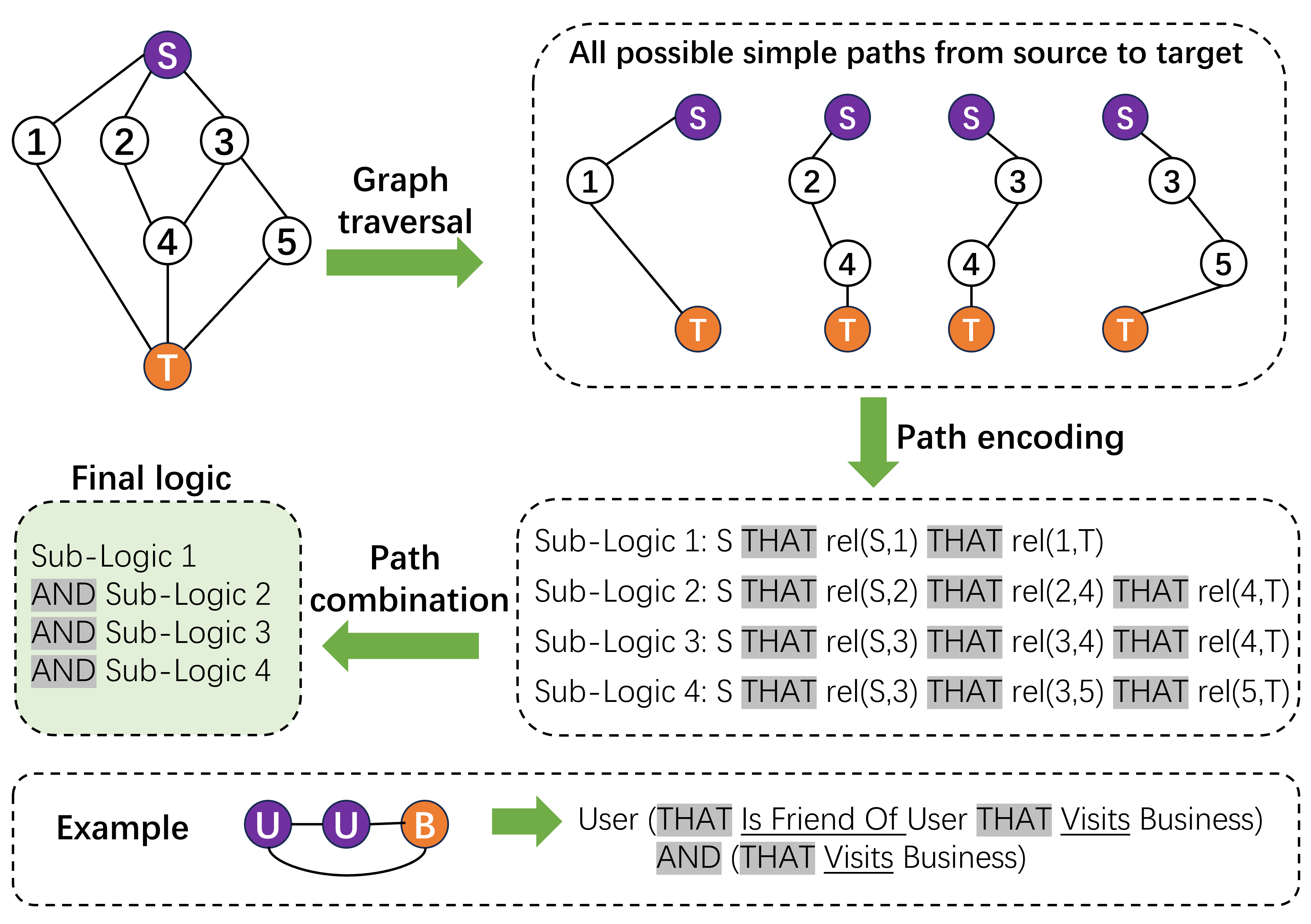}
    \caption{Natural language encoding of meta-structures.}
    \Description{Natural language encoding of meta-structures.}
    \label{fig:grammar}
\end{figure}

Previous works represent meta-structures either as matrices or sets of numbers~\cite{han2020genetic,ding2021diffmg}, which can be challenging to interpret in terms of their semantic meanings.
As a result, this poses obstacles for LLMs to effectively comprehend such representations. 
To address this limitation and enhance LLMs' comprehension of meta-structures, we design a \emph{grammar translator} module to encode each meta-structure into a natural language sentence, as shown in Figure~\ref{fig:grammar}.
For a given meta-structure, we begin by traversing its structure to find all possible simple paths connecting the source node to the target node.
Each resulting path is equivalent to a meta-path decomposed from the original meta-structure.
Next, we encode each path into a natural language sentence using nested clauses signified by a conjunction word \textit{THAT}, which is a commonly-used grammar in English and thus expected to be well-comprehended by the LLM.
In each clause, the central verb connecting two entities is the semantic meaning corresponding to the edge connecting two nodes. 
After obtaining the natural language encodings of the decomposed meta-paths, \textit{i.e.}, \textit{sub-logics}, we further combine them using another conjunction word \textit{AND} to convey the logical summation effect. 

\subsection{Basic Operations for Candidate Meta-Structure Generation} \label{method:basic_operations}
\begin{figure}[ht]
    \centering
    \includegraphics[width=\linewidth]{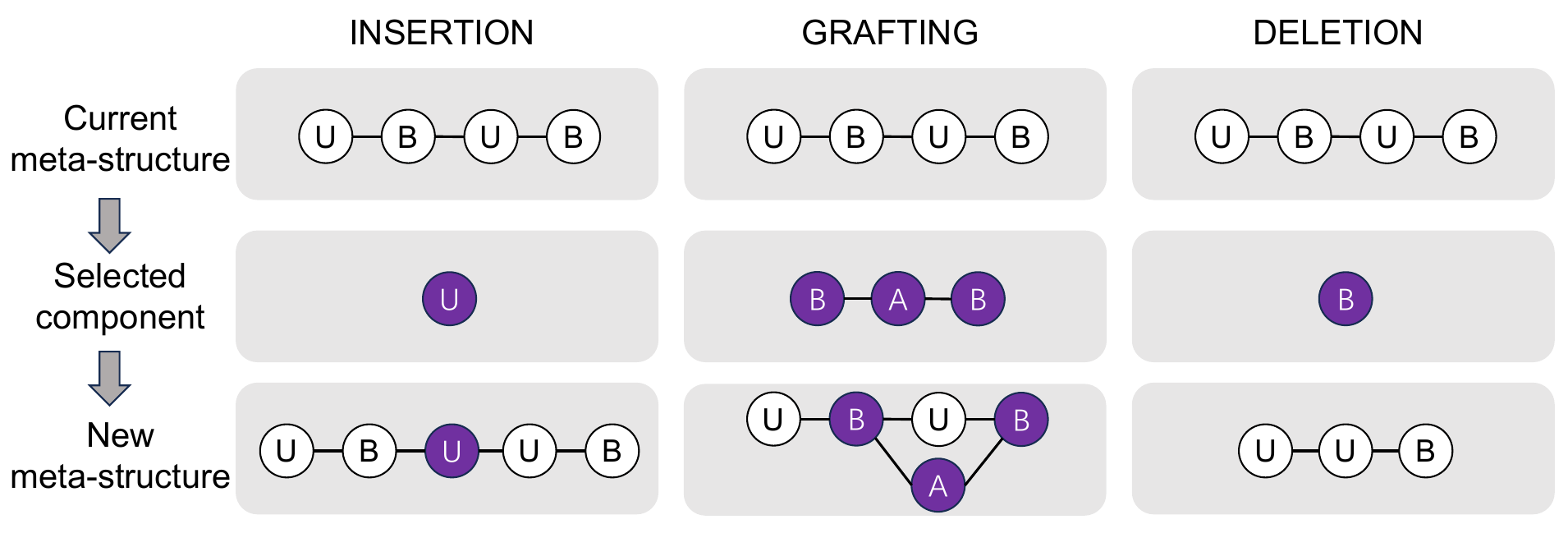}
    \caption{Basic operations of exploring adjacent meta-structures.}
    \Description{Basic operations of exploring adjacent meta-structures.}
    \label{fig:basic_operations}
\end{figure}

To generate comprehensive candidates for LLM selection while ensuring validity, we define three basic operations for modifying any meta-structure, and design a set of components for these operations, analogous to playing with Lego blocks.
Examples are shown in Figure~\ref{fig:basic_operations}.
\begin{itemize}
    \item \textit{INSERTION}. This operation replaces one edge of the original meta-structure with a component. It introduces new connections and expands the structure.
    \item \textit{GRAFTING}. This operation takes a component, finds two nodes in the original meta-structure with the same type as the chosen component's first and last node, and merges them respectively. It creates branching structures to enhance expressiveness.
    \item \textit{DELETION}. This operation removes a certain amount of nodes from the original meta-structure, and reconnects the remaining nodes to ensure the structure remains valid.
\end{itemize}
In our experiments, we take all meta-paths with no more than $2$ nodes as components for \textit{INSERTION}, and all meta-paths with no more than $3$ nodes as components for \textit{GRAFTING}.
\textit{DELETION} does not require components as input, as it regards all existing nodes on the original meta-structure as operation candidates.
While using components with more nodes expands the exploration space, it may also introduce complexity and confusion for the LLM. 
We leave it as future work to investigate the optimal component settings.

\subsection{Semantic Similarity-Oriented LLM Agents for Candidate Selection} \label{method:llm_selector}

To regulate the explainability of discovered meta-structures, we design two LLM agents that evaluate and select candidate meta-structures in a semantic similarity-oriented manner.
We illustrate the interaction processes between the main program and the LLM agents in Figure~\ref{fig:llm_prompt}.

\subsubsection{Few-Shot LLM Predictor}

After applying the basic operations on each meta-structure, we derive a set of one-step neighbors as potential candidates.
These neighbors can include meta-structures that have been encountered and evaluated in earlier generations, as well as entirely new ones. 
To leverage insights from previous evaluations and guide the decision-making process, we design an LLM agent as a \textit{few-shot LLM predictor} (abbreviated as ``predictor'' below).
This predictor estimates the performance $\hat{p}$ of each candidate through instruction tuning on a small set of structure-performance pairs sampled from a \textit{performance pool} that records meta-structures in all previous rounds.
Additionally, the predictor is asked to provide a self-estimated confidence value $\hat{c}$ for each prediction, resulting in a $(\hat{p}, \hat{c})$ pair associated with each candidate.
Intuitively, if the predictor considers a candidate to be highly similar to a counterpart in the \textit{performance pool}, it is likely to predict a similar performance and assign higher confidence to this prediction.
This is grounded in the understanding that structural similarity often implies functional similarity.
An illustrative prompt-response round is exemplified in Step 1 of Figure~\ref{fig:llm_prompt}. 

\subsubsection{Similarity-Oriented LLM Selector}

\begin{figure*}[ht]
    \centering
    \includegraphics[width=0.95\linewidth]{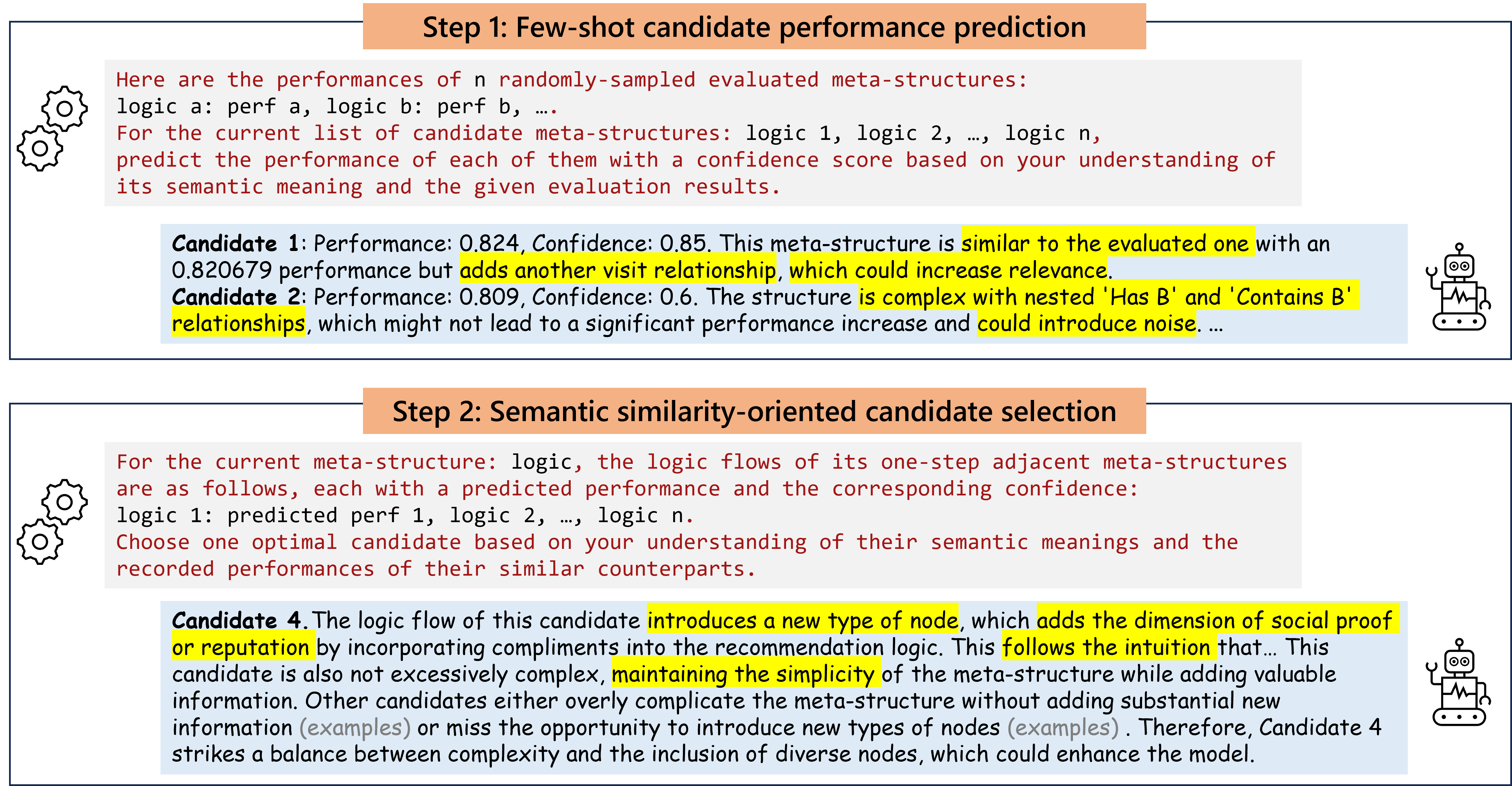}
    \caption{Example of LLM prompts and feedbacks.}
    \Description{Example of LLM prompts and feedbacks.}
    \label{fig:llm_prompt}
\end{figure*}

Upon receiving a set of candidates and their corresponding predicted performances from the \textit{few-shot LLM predictor}, we design another LLM agent, i.e., a \textit{similarity-oriented LLM selector} (abbreviated as ``selector'' below), to make the final decision of selecting one single candidate to proceed to the next generation.
During this process, the selector is expected to consider multiple factors simultaneously, and potentially trade-off between them in order to make the optimal decision (see Appendix~\ref{appendix:occam},\ref{appendix:tradeoff}).
These factors include: 
(1) Semantic meanings, which reflect the relevance and alignment of the meta-structure with the desired objectives and requirements.
(2) Structural complexities, which indicates the potential risks of overfitting.
(3) Expected outcomes provided by the \textit{few-shot LLM predictor}, which indicates the potential benefits from selecting a particular meta-structure in terms of performance improvement.
(4) Credibility of outcome expectation also provided by the \textit{few-shot LLM predictor}, which reflects the reliability and trustworthiness of the predictions.
An illustrative prompt-response round is exemplified in Step 2 of Figure~\ref{fig:llm_prompt}.

\subsection{Performance-Oriented Evolutionary Updater} \label{framework}

With closed-source LLM modules in the loop, it is not feasible for us to obtain the gradient for optimizing meta-structure search.
Therefore, we operationalize a derivative-free optimization framework with inspirations from the genetic algorithm.
Specifically, we maintain a population of $N$ individuals, each representing a distinct meta-structure.
In every generation, we first evaluate the performance of each meta-structure by using it to train a GNN for the given downstream task.
After evaluation, the underperforming meta-structures are \textit{eliminated} from the population.
The surviving meta-structures undergo a \textit{reproduction} phase, where duplication occurs with probabilities proportional to their performances.
In essence, this phase uses promising meta-structures to replenish the population to its original size.
Both \textit{elimination} and \textit{reproduction} processes mirror natural selection mechanisms that enable species evolution in the wild.
After getting the modified population, we feed it into the aforementioned LLM agents to for a new round of individual meta-structure improvement.
This step can be seen as a way of targeted \textit{mutation} within the evolutionary framework, as new nodes and/or edges can be generated and some of the existing nodes and/or edges may be removed.
The modified population will be re-evaluated at the onset of the next generation, forming a loop of derivative-free optimization.
In summary, by utilizing this evolutionary optimization framework, we can iteratively search for and improve meta-structures without relying on gradient-based optimization methods.

\subsection{Differential LLM Explainer Agent} \label{method:llm_explainer}

One prominent advantage of the LLM lies in its unparalleled ability for natural language generation.
To harness this ability, we design a \textit{differential LLM explainer} agent (abbreviated as ``explainer'' below) to automatically generate human-comprehensible textual explanations that elucidate the reasons behind the superior performance of discovered meta-structures.
To guide the explainer in discerning the critical structural properties that contribute to performance enhancement, we design a prompting process in the \textit{chain-of-thought} flavor~\cite{wei2022chain} .
Specifically, for analyzing a given meta-structure $T$, it unfolds in the following two steps:

\textit{Step 1: Structural Comprehension}. We begin by sampling a set of $n$ one-step neighbors for the meta-structure $T$, and translate each of them into a natural language sentence according to Method~\ref{grammar}. 
Then, we prompt the \textit{differential LLM explainer} to conduct a comprehensive analysis of both $T$ and all its sampled neighbors. 
This process involves breaking down each of them into meaningful sub-structures and identifying the functions os these sub-structures.

\textit{Step 2: Performance Attribution}. 
We first perform a quick evaluation of $T$ and each of the sampled neighbors separately by training a GNN with one structure at a time for downstream tasks. 
Then, we ask the \textit{differential LLM explainer} to identify the presence/absence of beneficial/detrimental sub-structures in the meta-structure $T$. 
This attribution process involves a joint consideration of the evaluated performances and the structural analysis conducted in the previous step.

The combination of these two steps empowers the explainer to unravel the intricate connections between structural properties and empirical performance, providing a comprehensive understanding of the discovered meta-structures.
The effectiveness of this module is further justified by a user study involving human evaluators, which is elaborated in Section~\ref{experiment:questionnaire}.

\section{Experiments}

\subsection{Experimental Settings}

\subsubsection{Datasets}
We evaluate \emph{ReStruct} on two important tasks in HIN learning: recommendation and node classification, each with four datasets covering different fields.
Detailed statistics of all eight datasets can be found in Appendix~\ref{appendix:dataset}. 

In the recommendation task, our goal is to predict the existence of links between source nodes (\textit{e.g.}, users) and target nodes (\textit{e.g.}, items or businesses).
We conduct experiments on four widely-used real-world datasets: Amazon, Yelp, Douban Movie (abbreviated as "Douban"), and LastFM\footnote{https://github.com/librahu/HIN-Datasets-for-Recommendation-and-Network-Embedding}.
For datasets including user ratings of items, the ratings are converted to 0-1 binary labels according to a threshold of $2$.
A label of 1 indicates the presence of preference, while 0 indicates the absence.
Among the user-item pairs with the label '1', we randomly select half of them as positive pairs, which are further randomly split into training-validation-testing sets with a ratio of 3:1:1.
The other half is reserved for network construction so as to prevent label leakage.
We take all user-item pairs with the label '0' as negative pairs, and also randomly split them into train-validation-test sets to pair each positive pair.
If the number of negative pairs is insufficient, we randomly sample unconnected items until reaching the desired number.
The evaluation metric used in these experiments is AUC (Area Under the ROC Curve), which measures the model's ability to rank positive instances higher than negative instances.

In the node classification task, our goal is to predict the labels of nodes belonging to a specific type, such as determining the genre of a movie. 
To evaluate the effectiveness of our approach, we perform experiments on four widely-adopted real-world datasets: ACM, IMDB, DBLP, and OAG-NN.
In these datasets, the classification targets correspond to the subjects of papers in ACM, the genres of movies in IMDB, the research areas of authors in DBLP, and the published venues of papers in OAG-NN, respectively.
For ACM, IMDB, and DBLP, we follow the data splits used in previous works~\cite{ding2021diffmg, li2023differentiable, yun2019graph}.
For OAG-NN~\cite{hu2020heterogeneous}, we filter the published venues with more than 100 recorded papers, and randomly split the dataset into training-validation-testing sets by 3:1:1.
The evaluation metric used in these experiments is the Macro-F1 score, which measures the performance of the classification model in terms of precision and recall.

\subsubsection{Baselines}
We compare \emph{ReStruct} with a set of state-of-the-art baselines.
These baselines can be classified into three categories:
\begin{itemize}
    \item Hand-crafted meta-paths: (1) metapath2vec~\cite{dong2017metapath2vec}, which trains a skip-gram model with meta-path guided random walks; (2) HIN2Vec~\cite{fu2017hin2vec}, which learns latent vectors by jointly training for multiple prediction tasks; (3) HAN~\cite{wang2019heterogeneous}, which is a heterogeneous GNN that learns graph representation with multiple hand-crafted meta-paths and fuses them with a multi-head attention mechanism; (4) HERec~\cite{shi2018heterogeneous}, which combines random walks with an extended matrix factorization model.
    \item Automatically-searched meta-paths: RMSHRec~\cite{ning2022automatic}, which adopts a reinforcement learning framework to search for meta-paths.
    \item Automatically-searched meta-structures: (1) GEMS~\cite{han2020genetic}, which employs a genetic algorithm; (2) DiffMG~\cite{ding2021diffmg}, which adopts a neural architecture search manner and searches for meta-structures in a differentiable manner; (3) PMMM~\cite{li2023differentiable}, which further generalizes DiffMG with multi-graph search.
\end{itemize}

\subsubsection{Hyperparameter Settings}
For our model, we run the algorithm for $30$ generations with a population size of $5$ and an elimination rate of $0.2$.
When modifying each meta-structure, we randomly sample a set of $20$ candidates if there are too many of them from the one-step neighbors.
When predicting meta-structure performances with the \textit{few-shot LLM predictor}, we randomly sample $30$ records from the \textit{performance pool} to fuel the few-shot learning paradigm. 
To implement the LLM agents, we use the GPT-4 model by calling the OpenAI API\footnote{https://platform.openai.com/docs/models/}, while robustness analysis with other LLM models are also carried out (see Section~\ref{section:robustness}).
We employ the DGL implementation of HAN\footnote{https://github.com/dmlc/dgl/tree/master/examples/pytorch/han}.
For all the other baseline models, we follow the implementation released by the authors.
We fix the hidden dimensions to $64$ for all evaluated models, and tune other hyperparameters including learning rate, weight decay, and dropout by referring to the performances on the validation set.
To reduce the noise brought by randomness during program execution, for each combination of (model, task, dataset), we run experiments with $10$ different random seeds and report the average performance with standard deviation.

\subsection{Comparison on Recommendation}

In Table~\ref{tab:recommendation_auc}, we report the experimental results of \emph{ReStruct} on the recommendation task compared to baselines.
First, we observe that models using meta-paths generally exhibit substantially lower performances than those using meta-structures, confirming that the stronger expression capability of meta-structures is desired for heterogeneous graph learning.
Second, \emph{ReStruct} consistently achieves the best performance across four datasets, showcasing the effectiveness of our framework in identifying meaningful and useful structures in various HINs.
In particular, the performance gain over GEMS confirms that the LLM-guided ``targeted mutation'' converges to better solutions than pure random mutation in a classic genetic framework.

\begin{table*}[ht]
\caption{AUC (\%) of recommendation on four datasets.}
\begin{tabular}{@{}p{1cm}p{1.5cm}p{1.1cm}p{1.1cm}llllllll@{}}
\toprule
\textbf{} & metapath2vec & HIN2Vec & HAN & RMSHRec & HERec & GEMS & DiffMG & PMMM & \textbf{\emph{ReStruct}}  \\ 
\midrule
Amazon & 56.88$\pm$0.27 & 58.66$\pm$0.12 & 60.03$\pm$0.48 & 61.80$\pm$0.15 & 70.55$\pm$0.05 & 63.76$\pm$0.47 & 73.27$\pm$0.20 & 73.78$\pm$0.04 & \textbf{75.27$\pm$0.19} \\
Yelp   & 52.29$\pm$0.59 & 68.06$\pm$2.31 & 63.88$\pm$2.32 & 59.39$\pm$0.81 & 68.37$\pm$0.34 & 75.46$\pm$0.28 & 77.63$\pm$0.40 & 76.76$\pm$0.17 & \textbf{84.04$\pm$0.24} \\
Douban & 52.84$\pm$0.00 & 85.95$\pm$0.06 & 63.27$\pm$0.38 & 79.72$\pm$0.32 & 92.22$\pm$0.00 & 90.84$\pm$0.10 & 93.94$\pm$0.04 & 94.31$\pm$0.02 & \textbf{94.49$\pm$0.03} \\ 
LastFM & 68.57$\pm$0.10 & 69.62$\pm$2.00 & 72.93$\pm$2.42 & 81.94$\pm$0.22 & 79.64$\pm$0.06 & 78.23$\pm$0.08 & 82.53$\pm$0.11 & 82.88$\pm$0.07 & \textbf{85.21$\pm$0.09} \\
\bottomrule
\end{tabular}
\label{tab:recommendation_auc}
\end{table*}

\begin{figure*}[ht]
    \centering
    \includegraphics[width=\linewidth]{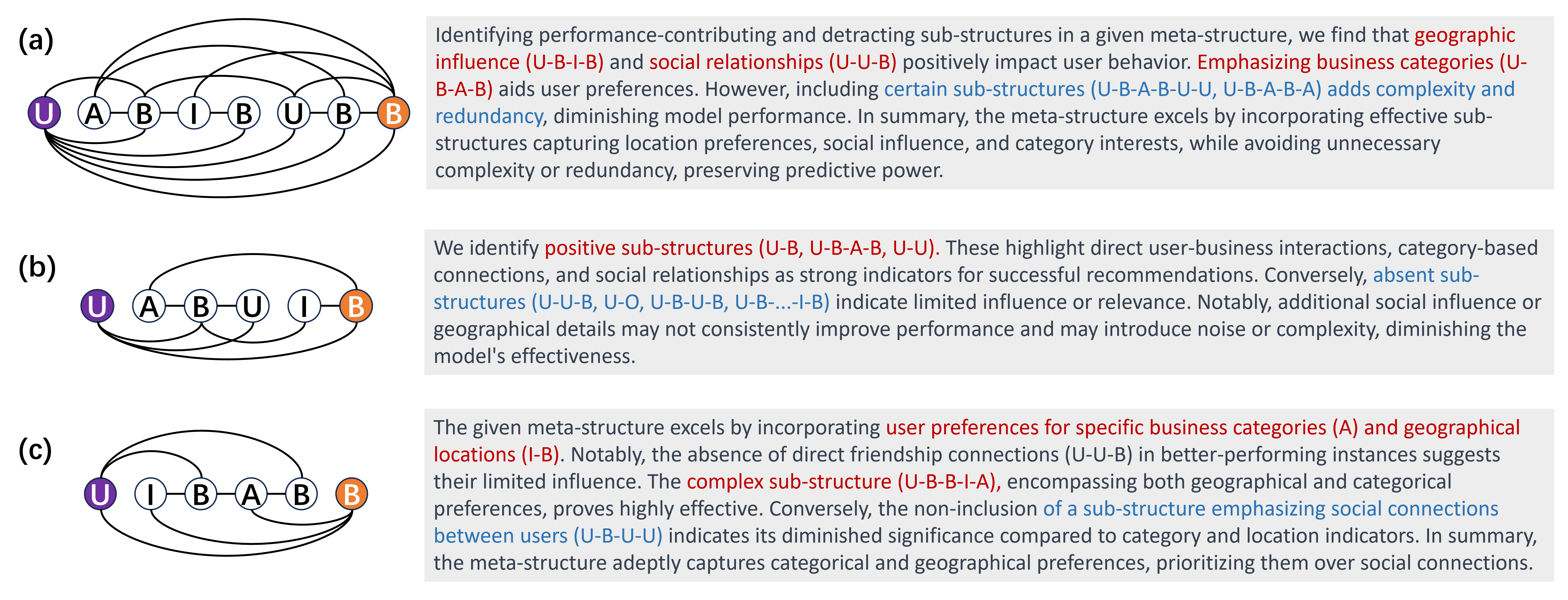}
    \caption{Discovered meta-structures on Yelp and the corresponding generated natural language explanations.}
    \label{fig:discovered_structures}
    \Description{Discovered meta-structures on Yelp and the corresponding generated natural language explanations.}
\end{figure*}

\subsection{Comparison on Node Classification}

In Table~\ref{tab:node_classification_f1}, we report the experimental results of \emph{ReStruct} on the node classification task compared to baselines.
First, metapath2vec, HIN2Vec, and HAN do not achieve desirable performances, mainly due to their heavy reliance on hand-crafted meta-structures.
Second, DiffMG and PMMM demonstrate improved performances, showcasing the advantage brought by meta-structures over meta-paths, as well as the NAS searching framework.
Finally, \emph{ReStruct} achieves the best performance on ACM, IMDB, and OAG datasets.
It closely aligns with state-of-the-art results on the DBLP dataset, where Macro F1 scores already exceed $94\%$ -- a level challenging to significantly surpass.

\begin{table*}[ht]
\caption{Macro F1 scores (\%) of node classification on four datasets.}
\begin{tabular}{@{}llllllll@{}}
\toprule
\textbf{} & metapath2vec & HIN2Vec & HAN & DiffMG & PMMM & \textbf{\emph{ReStruct}}  \\ 
\midrule
ACM     & 67.13$\pm$0.50 & 80.75$\pm$0.77 & 91.20$\pm$0.25 & 92.65$\pm$0.15 & 92.76$\pm$0.14 & \textbf{92.82$\pm$0.23} \\
IMDB    & 40.82$\pm$1.48 & 48.16$\pm$0.44 & 55.09$\pm$0.67 & 61.04$\pm$0.56 & 61.69$\pm$0.40 & \textbf{63.32$\pm$0.62} \\ 
DBLP    & 89.93$\pm$0.45 & 90.58$\pm$0.62 & 92.13$\pm$0.26 & 94.45$\pm$0.15 & 94.69$\pm$0.10 & \textbf{94.09$\pm$0.36} \\
OAG-NN  & 27.61$\pm$1.65 & 47.13$\pm$1.14 & 45.80$\pm$5.84 & 37.67$\pm$3.13 & 30.18$\pm$3.87 & \textbf{47.52$\pm$1.56} \\
\bottomrule
\end{tabular}
\label{tab:node_classification_f1}
\end{table*}

\subsection{Explainability Analysis}

\subsubsection{Visualization of Discovered Structures and LLM-generated Explanation}

Figure~\ref{fig:discovered_structures} showcases $3$ discovered meta-structures that are among the top-performing ones on the Yelp dataset for recommendation, each with a summary text explaining the structural attributes underlying their outstanding performances.
These summaries are generated by our \textit{differential LLM explainer} and further condensed for clarity.
We highlight the LLM-identified good (relevant) sub-structures in red, and the bad (distracting) sub-structures in blue.
This visualization facilitates a comprehensive understanding of each structure's composition.
For example, the meta-structure in Figure~\ref{fig:discovered_structures} (a) contains critical yet simple sub-structures describing the geographical, social, and business category contexts of user behavior. 
While more nodes and edges can create complex relationships, they are not always desirable for HIN learning, as showcased by the identified distracting sub-structures.
These structures, though challenging to handcraft, carry semantically meaningful explanations that remain accessible with textual assistance.

\subsubsection{User Study for Human Evaluation} \label{experiment:questionnaire}

We conduct a user study to evaluate the explainability provided by our framework compared to baselines from a human perspective. 
As our framework targets HIN researchers and engineers as potential users, we recruit $73$ graduate students with domain knowledge of HIN research as our participants.
To further ensure participants' solid understanding of the survey's processes and questions, we provide clear explanations of key concepts such as \textit{HIN}, \textit{meta-path} and \textit{meta-structure}, supported by illustrative examples at the beginning of the survey.
We structure the study around two sets of questions.
The first set of questions is designed to assess the inherent comprehensibility of generated meta-structures without textual explanations. 
To this end, we present the visualizations of the best meta-structures discovered by our model alongside those from three baseline models (GEMS, DiffMG, and PMMM)~\cite{han2020genetic, ding2021diffmg, li2023differentiable}, and ask the participants to select the most comprehensible one from their point of view.
The second set of questions is designed to assess the comprehension gain brought by the textual explanation generated by our \textit{differential LLM explainer} (see Method 3.5), coined the \textit{Differential Explanation}. 
As a baseline, we include a \textit{Non-Differential Explanation}, generated by directly prompting an LLM to explain the reasons behind a meta-structure's strong performance without undergoing the two-step prompting process.
For three meta-structures discovered by our model on the Yelp HIN, participants are presented with both types of explanations.
They are then asked to determine which one is more helpful in enhancing their understanding of how the meta-structure is constructed and gaining insights on how to design a better one.
By engaging participants in head-to-head comparisons, we aim to gather valuable feedback on the relative helpfulness of each explanation type. 
Before starting the survey, we also carried out a pilot study~\cite{rattray2007essential} with $5$ participants to ensure the clarity of the questions and visualizations, and none of them expressed confusion or difficulty in understanding these materials.
The complete questionnaire utilized in our study is available in our GitHub repository. 

\begin{figure}[ht]
    \centering
    \includegraphics[width=\linewidth]{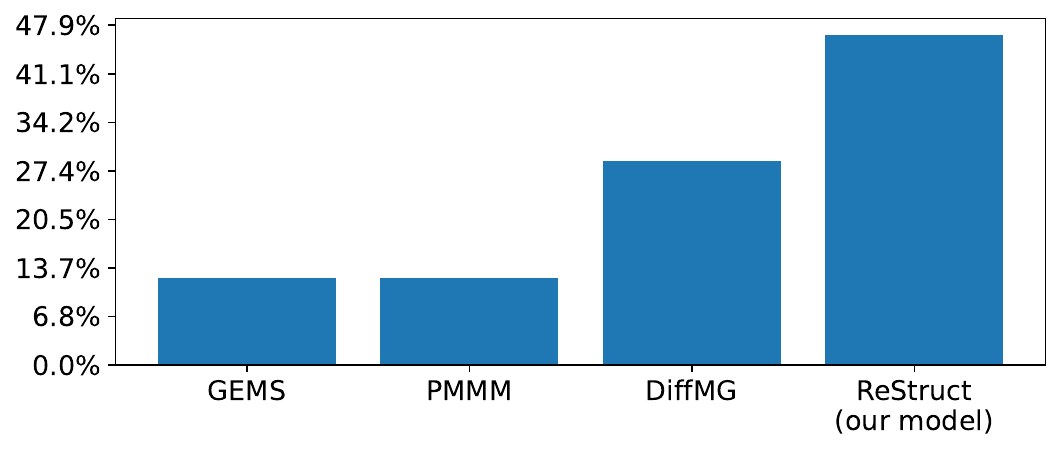}
    \caption{Human evaluated explainability of the best meta-structures discovered by different models.}
    \Description{Human evaluated explainability of the best meta-structures discovered by different models.}
    \label{fig:questionnaire_result_1}
\end{figure}

Figure~\ref{fig:questionnaire_result_1} shows the result of our first question set.
Among all four models, the meta-structure discovered by our model is regarded by most people ($46.6\%$) as the most informative or comprehensible one, outperforming the second baseline, DiffMG ($28.8\%$), by $61.8\%$.
GEMS and PMMM follow with the same level of recognition, $12.3\%$.
This outcome aligns with our initial objective of discovering meta-structures that not only excel in downstream tasks but are also accessible and easily understood by human users, presumably because our framework seeks to find meta-structures that are semantically meaningful while keeping as simple as possible.

\begin{figure}[ht]
    \centering
    \includegraphics[width=\linewidth]{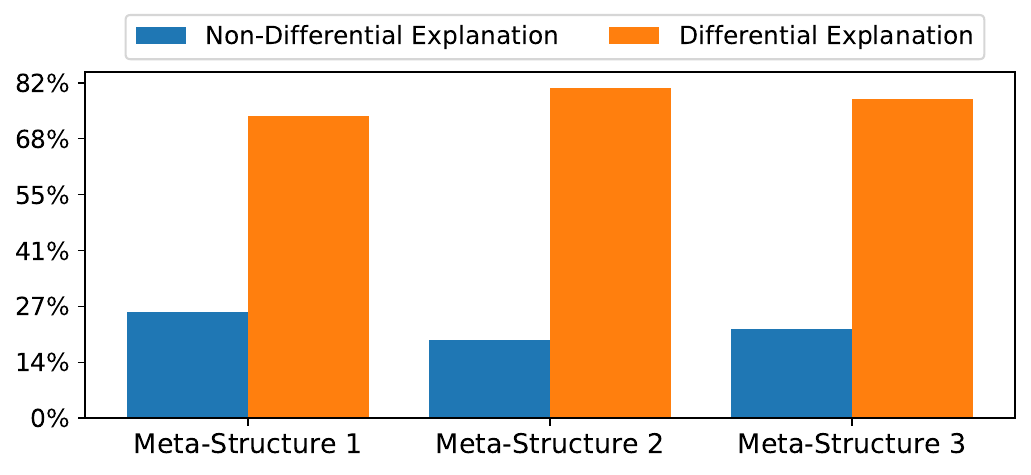}
    \caption{Explainability gain brought by LLM-generated differential meta-structure explanation compared to non-differential explanation.}
    \Description{Explainability gain brought by LLM-generated differential meta-structure explanation compared to non-differential explanation.}
    \label{fig:questionnaire_result_2}
\end{figure}

Figure~\ref{fig:questionnaire_result_2} shows the result of our second question set.
When comparing two types of generated textual explanations, the \textit{Differential Explanation} is consistently and significantly more preferred by human participants ($77.6\pm2.8\%$) across various meta-structures.
This outcome justifies the effective design of our \textit{differential LLM explainer} that unleashes the LLM's reasoning ability on the intricate connections between sub-structures, sub-functions, and how they interact to determine the ultimate performances.

\subsection{Robustness Analysis} \label{section:robustness}

To test the robustness of our method, we first analyze its performance variation under prompt perturbation. 
Following previous studies~\cite{sun2023evaluating,moraffah2024adversarial,zhang2024pptc}, we ask ChatGPT to paraphrase our prompts while maintaining the key module designs in the framework such as grammar translation and LLM-guided performance prediction, and use them to replace the original prompts in the \textit{ReStruct} framework. 
As shown in Table~\ref{tab:paraphrase_prompts}, we find that prompt paraphrasing does not have a clear impact on model performance, as long as the key designs are kept, the importance of which has been validated in the previous section.

\begin{table*}
    \caption{Model robustness to prompt perturbation with ChatGPT (taking recommendation on Yelp as an example).}
    \begin{tabular}{p{2cm}p{9.5cm}p{2cm}}
    \hline
    Index     &  Prompt for Paraphrasing & AUC (\%) \\ \hline
    0 (Original) & - & 84.04$\pm$0.24 \\ \hline
    1 & You are given an instruction. Now, paraphrase it into a new instruction with equivalent meaning. Instruction: \{original prompt\} & 84.02$\pm$0.10 \\ \hline 
    2 & You are provided with the utterance of a specific task and I need you to paraphrase it. The actual input, question, and examples in the task should not be changed. You should only paraphrase the instructions. Task: \{original prompt\} The paraphrased utterance:  & 84.01$\pm$0.26 \\ \hline 
    \end{tabular}
\label{tab:paraphrase_prompts}
\end{table*}

Second, we analyze \textit{ReStruct}'s robustness by changing the underlying LLM.
Specifically, we conduct experiments with nine different LLMs from three popular series (GPT, Mistral, and GLM), with different training data, training methods, and parameter sizes.
As shown in Table~\ref{tab:llm_version},  the model performances across these LLMs consistently and significantly outperform all baselines, demonstrating the robustness of our framework against LLM versions.

\begin{table}
    \caption{Model performance with different LLM versions (taking recommendation on Yelp as an example).}
    \begin{tabular}{p{5cm}p{2.5cm}}
    \hline
    Model & AUC (\%) \\ \hline
    gpt-4-1106-preview (original) & 84.04$\pm$0.24 \\ \hline
    gpt-3.5-turbo-1106& 84.03$\pm$0.11 \\ \hline
    mistral-tiny-2312& 83.65$\pm$0.17 \\ \hline
    mistral-small-2312& 83.89$\pm$0.09 \\ \hline
    mistral-small-2402& 83.96$\pm$0.12 \\ \hline
    mistral-medium-2312& 84.04$\pm$0.16 \\ \hline
    mistral-large-2402& 83.87$\pm$0.13 \\ \hline
    glm-3-turbo& 84.12$\pm$0.06 \\ \hline
    glm-4& 83.76$\pm$0.07 \\ \hline
    \end{tabular}
    \label{tab:llm_version}
\end{table}
\section{Related Works}

\subsection{Identifying Meta-structures on HINs}
Over the past decade, HINs have gained popularity for their ability to capture the complex relations between multi-typed nodes, which play important roles in various research areas such as information retrieval and social network modelling~\cite{shi2016survey}. 
Meta-path, a pre-defined path template of relation sequences, was proposed to measure the similarities between nodes on HIN~\cite{sun2011pathsim}. 
It allows search algorithms like \emph{PathSim}~\cite{sun2011pathsim} to find peer nodes that are connected by paths with different semantic meanings. 
The concept of meta-path was later extended beyond the linear relationship to a more general form of meta-structure~\cite{huang2016meta}, where the relation patterns between connected nodes can be characterized as a directed acyclic graph. 
Previous works have found meta-structures useful for boosting machine learning performance on HINs~\cite{jiang2017semi}. 
However, early works were based on carefully hand-crafted meta-structures, which heavily relied on experts' domain knowledge. 
To address this challenge, several recent works proposed to automate meta-structure design with heuristic algorithms~\cite{meng2015discovering}, reinforcement learning~\cite{yang2019similarity}, evolutionary search~\cite{han2020genetic} and differentiable structure learning~\cite{ding2021diffmg}.
Different from previous efforts of automatic meta-structure design, we are first to leverage the emergent reasoning ability of LLMs~\cite{wei2022emergent} for this task. 
We design novel LLM agents for the automatic generation, evaluation, and explanation of novel meta-structures, which are proven effective in eight representative datasets.

\subsection{Deep Learning on HINs}

The success of graph neural networks introduces revolutionary deep learning techniques into HIN modeling~\cite{wu2020comprehensive}. 
Metapath2Vec~\cite{dong2017metapath2vec} proposed to learn deep representations for nodes via meta-path-guided random walks. 
Attention mechanism was later introduced to learn more expressive representations for HINs, proving effective for link prediction~\cite{xu2019relation} and node classification~\cite{wang2019heterogeneous}. 
Subsequent research endeavors focused on designing more effective heterogeneous GNN frameworks~\cite{li2021leveraging,fu2020magnn}.
Besides, considerable research efforts were drawn to replace handcrafted meta-structures with automatic search. 
GEMS~\cite{han2020genetic} proposed to combine heterogeneous GNN with evolutionary algorithms to identify meta-structures and learn deep neural networks simultaneously. 
Several deep reinforcement learning models and neural architecture search models are also proposed to jointly optimize the meta-structures with heterogeneous GNN~\cite{ding2021diffmg,peng2021reinforced,ning2022automatic,li2023differentiable,wan2020reinforcement,li2021graphmse}. 
However, previous automatic meta-structure design method solely focused on prediction performance, often yielding complex and difficult to explain meta-structures. 
To the best of our knowledge, our study is the first to harness the semantic reasoning capability of LLMs for automatic meta-structure design. 
Our model can discover meta-structures that not only show high prediction performance, but also can be adequately explained with natural languages.

\subsection{LLM for Graph Learning}

LLMs have demonstrated general capabilities beyond natural language tasks, attracting graph learning researchers who are particularly interested in leveraging their ability for graph reasoning tasks~\cite{guo2023gpt4graph}. 
Several attempts have been made to enhance node features using LLMs or employ them as standalone graph predictors~\cite{chen2023exploring}. 
The research community is also actively discussing the perspective of developing large graph models~\cite{zhang2023graph,liu2023towards}.
Recent research has found that LLMs exhibit certain ability for graph tasks such as detecting connectivity and cycles, performing topological sort, and emulating GNNs~\cite{wang2023can}.
Besides, LLMs can effectively perform reasoning on knowledge graphs~\cite{sun2023think}.
To better align graph problems with LLMs, recent works propose various methods to encode the geometric structure and node features of graph problems~\cite{zhao2023graphtext,fatemi2023talk}. 
With these encodings, researchers have explored the possibility of replacing GNNs with LLM reasoning~\cite{ye2023natural} and performing instruction tuning for graph tasks~\cite{tang2023graphgpt}. 
In this paper, we fundamentally extend LLM reasoning to HIN meta-structure discovery. 
Specifically, we propose a novel meta-structure encoding method, which effectively boosts LLMs' reasoning capability on HINs.

\subsection{LLM for Pattern Discoveries}
The scaled-up language models have emerged reasoning capability for general tasks~\cite{wei2022chain}, including commonsense reasoning, logical reasoning, and mathematics reasoning.
With the help of optimized prompting routines such as \emph{chain-of-thought} (CoT)~\cite{wei2022chain} prompting and \emph{tree-of-thoughts} (ToT)~\cite{yao2023tree} prompting, the scaling curve of LLMs' reasoning capability can be further effectively improved.
As a result, recent research has shown that LLMs can be leveraged to identify novel patterns and feasible solutions in large problem spaces. 
For example, \emph{FunSearch} was proposed to discover algorithm programs for solving mathematical problems~\cite{romera2023mathematical}.
Previous works also designed LLM-driven algorithms for evolutionary search~\cite{guo2023connecting}, reinforcement learning~\cite{shinn2023reflexion}, and hyper-parameter optimization~\cite{liu2024large}. 
In this paper, we propose a novel framework to harness the reasoning power of LLM for meta-structure discovery. 
Our framework equips LLMs with enhanced capability to understand the semantic meaning of meta-structures and search for promising candidates. 

\section{Conclusion}

In this work, we propose a novel framework, \emph{ReStruct}, that fuses the power of LLMs with evolutionary algorithms to facilitate automatic meta-structure discovery across diverse HINs.
On both recommendation and node classification tasks, extensive experiments demonstrate that \emph{ReStruct} excels in uncovering previously undiscovered meta-structures, thereby significantly enhancing downstream model performance compared to a set of state-of-the-art baselines.
Notably, a user study involving human participants confirms that \emph{ReStruct} substantially outperforms baseline methods in terms of the comprehensibility of discovered meta-structures and usefulness of generated explanations.
For future work, we will explore the feasibility of finetuning local models to mitigate network communication costs associated with API calls.

\section{Acknowledgement}
This work was supported in part by the National Key Research and Development Program of China under Grant 23IAA02114, the Guangzhou Municipal Nansha District Science and Technology Bureau under Contract No.2022ZD012, and the National Natural Science Foundation of China under Grant U23B2030 and Grant U22B2057.

\newpage
\bibliographystyle{ACM-Reference-Format}
\bibliography{reference}
\clearpage
\appendix

\section{Statistics of Datasets} \label{appendix:dataset}

\begin{table}[h]
\centering
\caption{Statistics of datasets for node classification.}
\label{tab:datasets_for_node_classification}
\resizebox{\columnwidth}{!}{
\begin{tabular}{lcccc}
\hline
Dataset             & ACM                                                                          & IMDB                                                                         & DBLP                                                                            & OAG-NN                                                                                       \\ \hline
\begin{tabular}[c]{@{}l@{}}Node \\ types\end{tabular}    & \begin{tabular}[c]{@{}c@{}}Author (A)\\ Paper (P)\\ Subject (S)\end{tabular} & \begin{tabular}[c]{@{}c@{}}Movie (M)\\ Actor (A)\\ Director (D)\end{tabular} & \begin{tabular}[c]{@{}c@{}}Author (A)\\ Paper (P)\\ Conference (C)\end{tabular} & \begin{tabular}[c]{@{}c@{}}Paper (P)\\ Author (A)\\ Affiliation (I)\\ Field (F)\end{tabular} \\ \hline
\begin{tabular}[c]{@{}l@{}}Edge \\ types\end{tabular}    & \begin{tabular}[c]{@{}c@{}}A-P, P-A,\\ P-S, S-P\end{tabular}                 & \begin{tabular}[c]{@{}c@{}}M-D, D-M,\\ M-A, A-M\end{tabular}                 & \begin{tabular}[c]{@{}c@{}}A-P, P-A,\\ P-C, C-P\end{tabular}                    & \begin{tabular}[c]{@{}c@{}}P-P,\\ P-A, A-P,\\ P-F, F-P,\\ A-I, I-A\end{tabular}                \\ \hline
\# Nodes      & 8,994                                                                        & 12,624                                                                       & 18,405                                                                          & 64,203                                                                                       \\ \hline
\# Edges      & 25,922                                                                       & 37,288                                                                       & 67,946                                                                          & 403,974                                                                                      \\ \hline
\# Classes    & 3                                                                            & 3                                                                            & 4                                                                               & 8                                                                                            \\ \hline
\# Training   & 600                                                                          & 300                                                                          & 800                                                                             & 2,334                                                                                        \\ \hline
\# Validation & 300                                                                          & 300                                                                          & 400                                                                             & 778                                                                                          \\ \hline
\# Testing    & 2,125                                                                        & 2,339                                                                        & 2,857                                                                            & 778                                                                                          \\ \hline
\end{tabular}}
\end{table}

\begin{table}[h]
\centering
\caption{Statistics of datasets for recommendation.}
\label{tab:datasets_for_recommendation}
\resizebox{0.9\columnwidth}{!}{
\begin{tabular}{@{}ccccc@{}}
\toprule
Dataset                 & Relations (S-T)              & \# S            & \# T            & \# S-T           \\ \midrule
\multirow{5}{*}{Yelp}   & \textbf{User-Business (U-B)} & \textbf{16,239} & \textbf{14,284} & \textbf{84,993}  \\
                        & User-User (U-U)              & 16,239          & 16,239          & 158,590           \\
                        & User-Compliment (U-O)        & 16,239          & 11              & 76,875            \\
                        & Business-Category (B-A)      & 14,284          & 511             & 40,009            \\
                        & Business-City (B-I)          & 14,284          & 47              & 14,267            \\ \midrule
\multirow{6}{*}{Douban} & \textbf{User-Movie (U-M)}    & \textbf{13,367} & \textbf{12,677} & \textbf{500,515} \\
                        & User-User (U-U)              & 13,367          & 13,367          & 4,085             \\
                        & User-Group (U-G)             & 13,367          & 2,753           & 570,047           \\
                        & Movie-Actor (M-A)            & 12,677          & 6,311           & 33,587            \\
                        & Movie-Director (M-D)         & 12,677          & 2,449           & 11,276            \\
                        & Movie-Type (M-T)             & 1,2677          & 38              & 27,668            \\ \midrule
\multirow{4}{*}{Amazon} & \textbf{User-Item (U-I)}     & \textbf{6,170}  & \textbf{2,753}  & \textbf{86,191}  \\
                        & Item-View (I-V)              & 2,753           & 3,857           & 5,694             \\
                        & Item-Category (I-C)          & 2,753           & 22              & 5,508             \\
                        & Item-Brand (I-B)             & 2,753           & 334             & 2,753             \\ \midrule
\multirow{3}{*}{LastFM} & \textbf{User-Artist (U-A)}   & \textbf{1,892}  & \textbf{17,632} & \textbf{46,417}   \\
                        & User-User (U-U)              & 1,892           & 1,892           & 25,434            \\
                        & Artist-Tag (A-T)             & 17,632          & 9,718           & 108,437            \\ \bottomrule
\end{tabular}}
\end{table}

\section{Emergence of Occam's Razor Phenomenon} \label{appendix:occam}

\begin{figure}[ht]
    \centering
    \includegraphics[width=0.85\columnwidth]{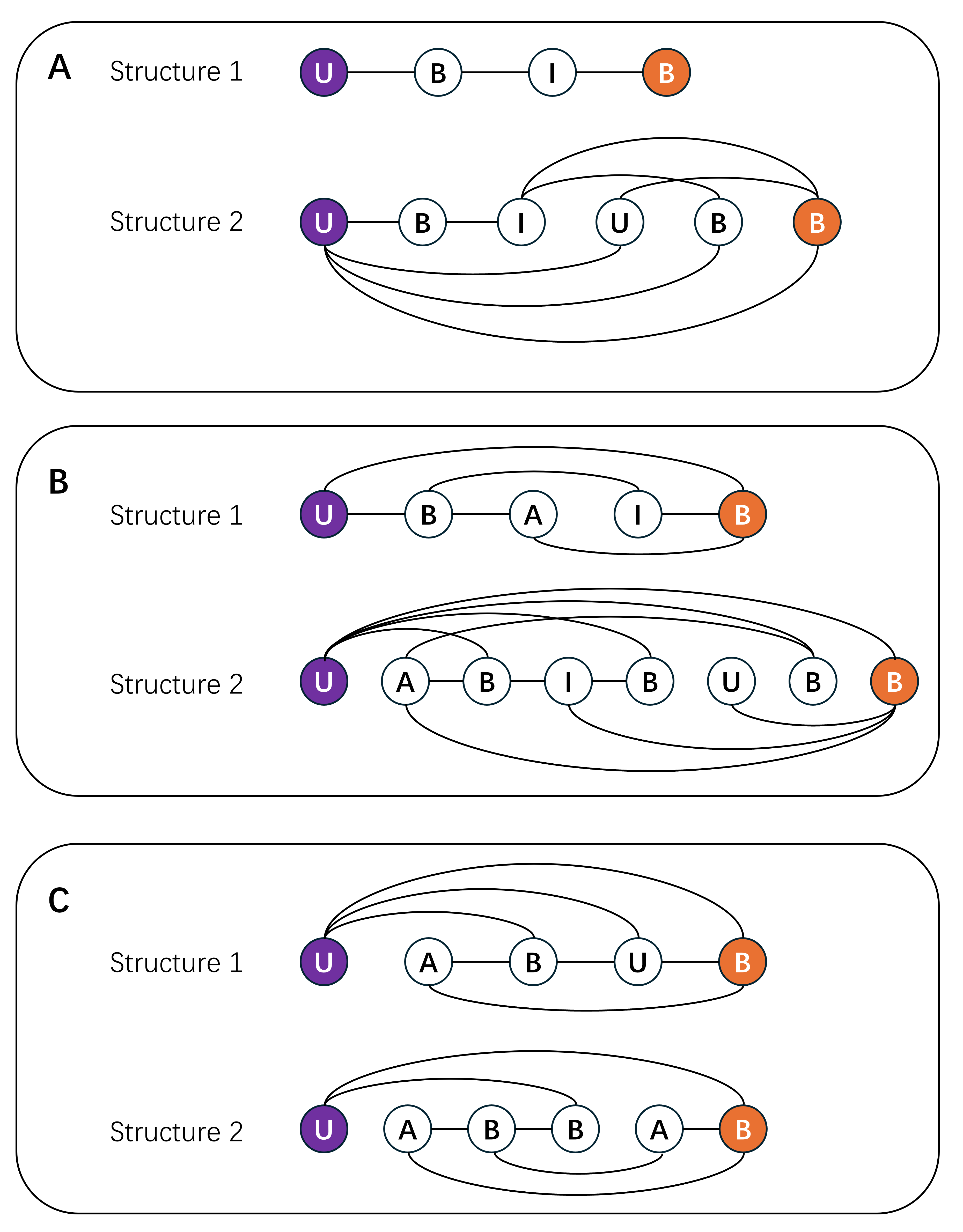}
    \caption{Visualization of meta-structure pairs for comparison.}
    \Description{Visualization of meta-structure pairs for comparison.}
    \label{fig:overfitting_pairs}
\end{figure}

In our user study, we demonstrate that \textit{ReStruct} finds meta-structures that are more comprehensible to human researchers (Figure~\ref{fig:questionnaire_result_1}), implying a tendency to avoid overcomplicated meta-structures. 
To provide further evidence, we add an experiment on the Yelp recommendation task, asking the LLM to select between meta-structure pairs with near-equal performance on the validation set but varying structural complexity (\#nodes and \#edges), which are visualized in Figure~\ref{fig:overfitting_pairs}.
As shown in Table~\ref{tab:prevent_overfit}, LLM consistently prefers meta-structures with lower structural complexity, citing reasons such as "simplicity," "generalizability," and even invoking "Occam's razor". 
Therefore, LLMs possess human-like preferences for simpler and more comprehensible meta-structures that mitigates overfitting.

\begin{table*}[ht]
\caption{LLM selection between meta-structure pairs with near-equal validation performances: three examples.} 
\begin{tabular}{p{1.3cm}p{1.8cm}p{14cm}}
\hline
Pair & Chosen Index & Reasons \\ \hline
1 (Fig.~\ref{fig:overfitting_pairs}A) & 1 & ``- Complexity vs. Performance: The second meta-structure has a more complex path, suggesting potentially richer semantics being captured. However, if both meta-structures show almost equal performance, the additional complexity might not translate into a significant performance gain. In such cases, it is generally better to opt for the simpler model (Occam's razor), which is why Meta-structure 1 is preferred.
- Overfitting Risks: The more complex second meta-structure might be more prone to overfitting, as it captures a more specific set of interactions. The simpler Meta-structure 1 might generalize better to unseen data.
- Interpretability: Meta-structure 1 is easier to interpret because it has a shorter path and fewer node interactions. This can be beneficial for understanding why certain recommendations are being made, which is important for transparency and trust in the recommendation system.
- Computational Efficiency: With shorter paths and fewer relations, Meta-structure 1 is likely to be more computationally efficient for both training and inference, which can be important in practical, large-scale applications.
- Meta-path Relevance: The paths in Meta-structure 1 adequately capture the relevant user-business interactions via a business's location. Adding more nodes and relations in the second meta-structure might not contribute much to capturing the essential features for the recommendation task.
In conclusion, the first meta-structure is selected as it achieves a good balance between simplicity, performance, and interpretable results while potentially being more computationally efficient and generalizable.” \\ \hline 
2 (Fig.~\ref{fig:overfitting_pairs}B) & 1 & ``- Path Simplicity: Meta-structure 1 offers a more straightforward path from 'U' (User) to 'B' (Local Business) through 'A' (Business Category) and 'I' (Business City), with minimal backtracking or redundant connections. The path follows a logical progression from user preference to business through the context of category and location.
- Computational Efficiency: The simpler meta-structure 1 should, in theory, be computationally less intensive because it involves fewer steps and connections. This could be advantageous when scaling the recommendation system and processing a large dataset.
- Interpretability: Meta-structure 1 is easier to interpret. In the context of a recommendation system, a clear logical flow from user to business via category and city is likely to align with a user's thought process when seeking recommendations, enhancing the user experience.
- Relevance of Categories and Location: By going through 'A' (Business Category) and 'I' (Business City), Meta-structure 1 reflects important aspects of businesses that users consider while making choices—what kind of service or product a business offers (category) and where it is located (city).
While Meta-structure 2 includes a potentially rich and complex connection pattern that might capture more nuanced relationships, the balance between complexity and utility must be considered, particularly if both structures have shown similar performance on the validation set. Given these considerations, the simpler, more interpretable, and likely more efficient Meta-structure 1 would likely be the optimal choice for the recommendation model." \\ \hline
3 (Fig.~\ref{fig:overfitting_pairs}C) & 1 & ``- **Simplicity and Interpretability**: The first meta-structure is less complex as it contains a shorter sequence of nodes, potentially making it easier to interpret and explain how the recommendation is generated. A simpler model is often preferred when performances are almost equal because it can be more transparent, and it can generalize better, avoiding overfitting the data.
- **Computational Efficiency**: Fewer node hops suggest lower computational costs in terms of executing the meta-path. This can be important for scaling the recommendation model to a large number of users or in an environment where computational resources are constrained.
- **Redundancy Avoidance**: The second meta-structure involves a repeated sequence of 'Business Category' to 'Local Business' ('A' to 'B') hops, which may introduce redundancy that does not necessarily contribute to predictive performance. Such over-specified meta-structures can sometimes lead to overfitting, where the performance on the validation set is good, but the generalization to unseen data may suffer.
- **Logical Coherence**: The first meta-structure provides a more direct flow of information from 'User' to 'Local Business' through their categories and mutual connections to other users. This logical flow is more coherent compared to the second meta-structure, which seems to loop back through 'Business Category' without a clear additional benefit.
- **Real-World Significance**: From a real-world perspective, the first meta-structure captures a user's preference through the categories of businesses they have visited and the influence of their friends' preferences for the same business categories. This could be a robust basis for recommendations without the need for the extra category-business-category loop in the second meta-structure.
Given these reasons, meta-structure 1 seems to be the optimal choice for improving the recommendation model on this Yelp HIN." \\ \hline
\end{tabular}
\label{tab:prevent_overfit}
\end{table*}

\section{Evidence of opportunity-risk tradeoff during candidate selection} \label{appendix:tradeoff}

\begin{figure}[ht]
    \centering
    \includegraphics[width=\columnwidth]{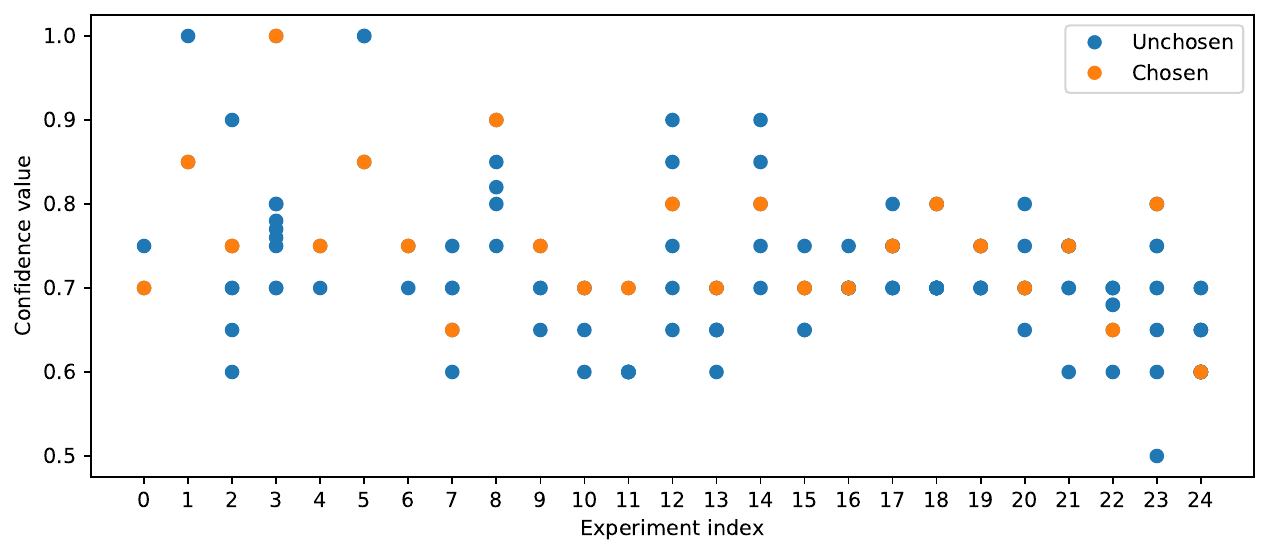}
    \caption{Comparison of confidence values between chosen and unchosen candidates.}
    \Description{Comparison of confidence values between chosen and unchosen candidates.}
    \label{fig:confidence-value}
\end{figure}

To check whether there is an overreliance on confidence values during candidate selection, we add an experiment to compare the confidence values between chosen and unchosen meta-structure candidates. 
As shown in Figure~\ref{fig:confidence-value}, \textit{ReStruct} does not adhere strictly to the candidates with the highest confidence values. 
Take Experiment \#14 as an example: although Candidate $0$ and $5$ have higher confidence values, \textit{ReStruct} decides to choose Candidate $3$, likely due to its higher predicted performance via effective semantic analysis. 
This highlights \textit{ReStruct}'s attempts to balance between opportunities and risks, when predicted performances and confidence values become trade-offs.

\end{document}